\pdfoutput=1
\documentclass[11pt,a4paper]{mvigreport}

\usepackage[numbers,sort&compress]{natbib}
\usepackage{marvosym}
\usepackage{pgfplots}
\pgfplotsset{compat=1.18}
\hypersetup{pdftitle={UniFlowWAM: A Unified World--Action Model with Visual Semantics and 4D Geometry}}

\newcommand{\tabnote}[1]{\par\vspace{3pt}\noindent\begin{minipage}{\linewidth}\scriptsize #1\end{minipage}}

\newlength{\figw}
\AtBeginDocument{\setlength{\figw}{0.65\textwidth}}
\newcommand{\figpanelsep}{\hspace{3em}}

\title{UniFlowWAM: A Unified World--Action Model \\ with Visual Semantics and 4D Geometry}
\shorttitle{UniFlowWAM}
\makeatletter
\newcommand{\mvigwordmarksmall}{\mvig@vc{\sffamily\fontsize{13}{13}\selectfont
  {\fontseries{eb}\selectfont\color{MVIGInk}MVIG}{\fontseries{m}\selectfont\color{MVIGGray}@SJTU}}}
\makeatother
\setlogos{\mvigwordmarksmall\logosep\siilogo\logosep\headerlogo{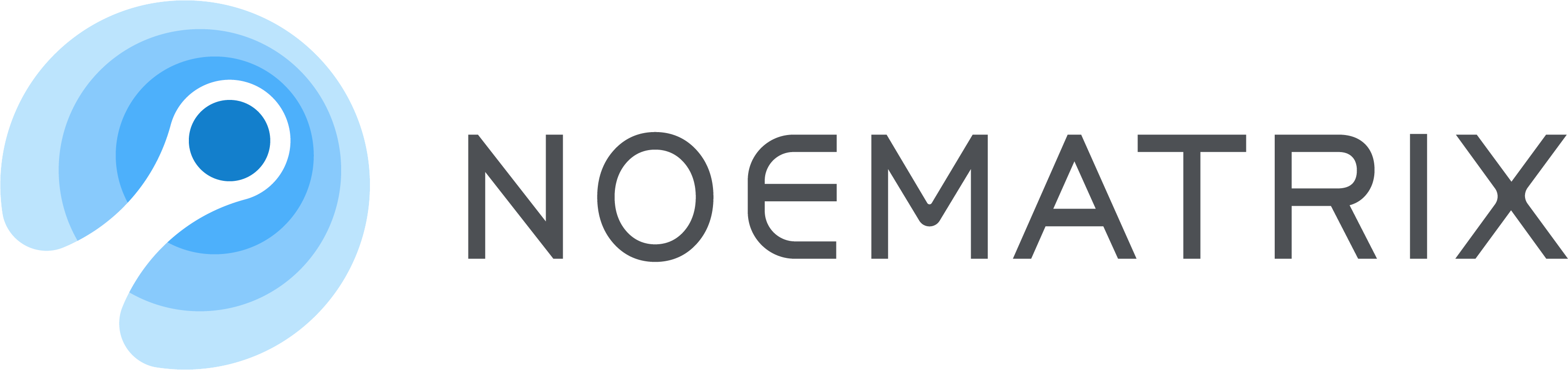}}
\reportlabel{Research Paper}
\teamname{}

\author[1,2,*]{Jialin Tian}
\author[1,3,*]{Zhuofan Li}
\author[2,3]{Xinkai Wang}
\author[1,2,3]{Bokai Lin}
\author[2]{Chenyi Wang}
\author[1]{Xinyu Zhan}
\author[1,2,3]{Cewu Lu}
\author[{\protect\Letter}1,2,3]{Lixin Yang}
\affil[1]{MVIG@SJTU}
\affil[2]{Shanghai Innovation Institute}
\affil[3]{Noematrix}
\reportnote{$^{*}$Equal contribution.\quad \textsuperscript{\Letter}Corresponding author.}
\projectpage{https://kksham111.github.io/UniFlowWAM_ProjectPage/}

\keywords{}

\begin{document}

\begin{abstract}
Robot demonstrations record what the robot saw and what it did.
They do not specify how the 3D world changed in between.
Scene-level 3D flow describes it by tracking 3D scene points, providing the world representation with 4D geometry.
Prior work has used flow to improve robot policies.
Flow served as an intermediate prediction that a separate model converts
into actions, as a prior injected at inference time, or as an extra output
learned together with the action.
In all of these designs, flow stays outside the policy, as a separate module
or as an extra prediction head.
We introduce UniFlowWAM, a Unified World--Action Model that makes 3D scene flow a
third modality on par with vision and action.
Its world representation therefore covers both visual semantics, through future
visual features, and 4D geometry, through 3D flow, and flow receives its own
supervision.
UniFlowWAM is a diffusion transformer with one expert per modality, and the three
experts exchange information through joint attention.
We train UniFlowWAM under a ten-objective role grid, where each objective assigns
every modality exactly one of three roles, namely target, condition, or absent.
The same model thus serves as a policy, as forward and inverse dynamics, and as a
predictor of future visual features or 3D flow.
We evaluate UniFlowWAM on RoboCasa-GR1, LIBERO, and LIBERO-Plus in simulation, and
on four real-world bimanual manipulation tasks.
On RoboCasa-GR1, UniFlowWAM raises the success rate of its base model without flow
from 52.1\% to 56.4\%.
It preserves in-distribution success on LIBERO at 98.1\% and reaches 87.7\%
zero-shot on LIBERO-Plus.
On the real robot, it achieves 87.0\% average success, compared with 68.5\%
for the strongest baseline.
\end{abstract}

\maketitle
\begin{figure}[!htb]
\centering
\includegraphics[width=\textwidth]{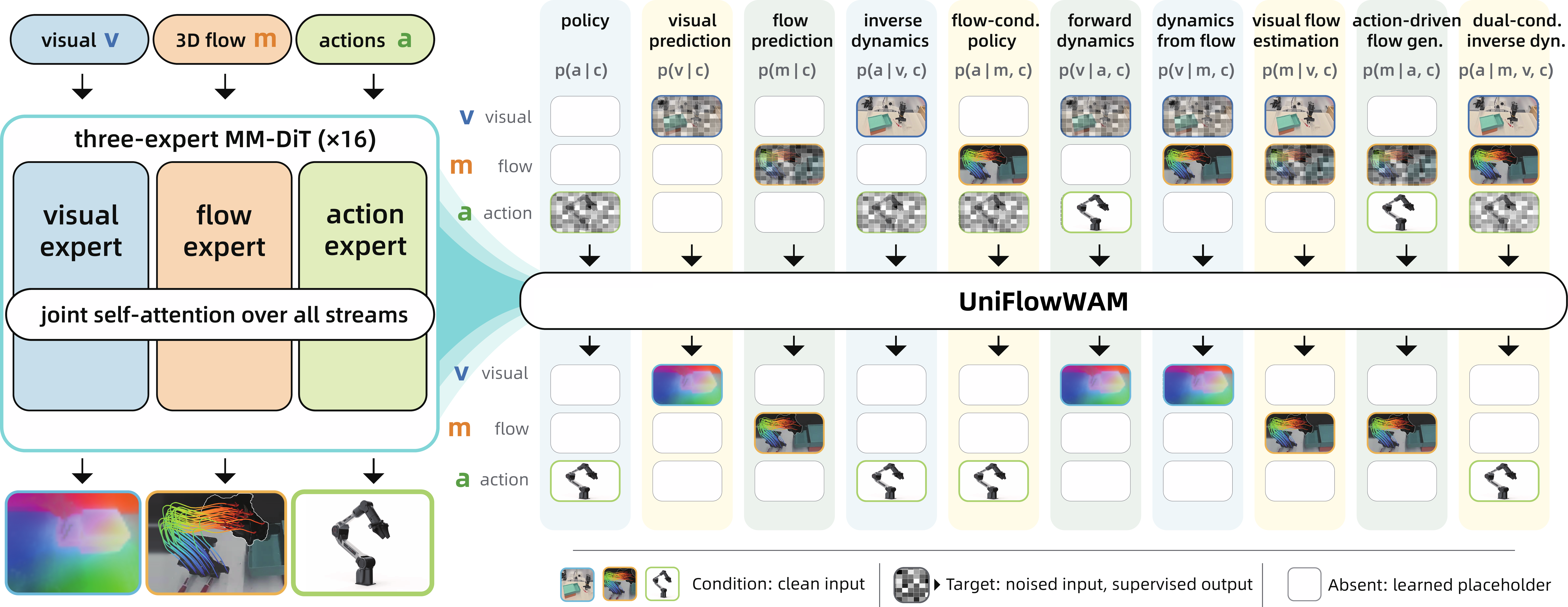}
\caption{\textbf{UniFlowWAM overview}: UniFlowWAM couples visual features, scene-level 3D flow, and actions in one three-expert diffusion transformer, denoising the three jointly so that 3D flow carries scene displacement into the space where actions are executed. A grid of ten objectives assigns each modality the role of target, condition, or absent, so a single model covers policy, dynamics, and prediction, yet deploys as a direct policy that needs no flow at test time.}
\label{fig:teaser}
\end{figure}

\section{Introduction}

A robot manipulates objects by moving its own body. Its action labels only record the commands that its joints or end-effector execute. What those commands do to the scene is left unwritten. A mug pushed across the table and a mug lifted and set back down can end in the same pose. The action labels do not describe how the mug translates or rotates along the way. Among possible world representations in a world--action model, a goal image describes the resulting state, but not the transition. A video sequence captures this process visually, but does not describe how scene points move in 3D. Scene-level 3D flow makes the effects of robot commands on the scene explicit: it describes how each scene point is displaced in 3D over time, providing a 4D geometric representation of this process.

Prior work has used scene-level flow to improve robot policies. Flow served as intermediate representations that were predicted first and then converted into actions in a separate stage~\cite{im2flow2act,generalflow}. A learned 3D scene-flow prior sharpened a vision-language-action policy when injected at inference~\cite{lamp}. Concurrent work trains sparse 3D keypoint flow jointly with the action~\cite{chronoflow}. These results establish flow as a useful control signal and as a variable that can be learned with actions.

Unified future-image--action models learn actions together with predictions of the future. UWM~\cite{uwm} and UVA~\cite{uva} train the policy together with forward and inverse dynamics and visual forecasting, so that both action-labeled and actionless data contribute. Latent-dynamics variants adopt the features of a frozen visual encoder~\cite{dinov3} as the future-state target and scale to large robotic datasets~\cite{lda}. The same model learns both to anticipate the visual consequences of actions and to infer actions from future visual features.
Those features capture the semantics in the future goal image, but do not explicitly describe how the 3D scene evolves to reach that state.
We ask: \emph{can a pretrained visual--action model learn where the 3D points of the scene go, and, more importantly, does learning this improve the actions it predicts?}

We explicitly supervise the 3D displacement of scene points, beyond the geometric state that depth-based targets supervise.
We introduce \textbf{UniFlowWAM}, a Unified World--Action Model whose world representation joins future visual semantics from DINOv3~\cite{dinov3} with 4D geometry from scene-level 3D flow.
Actions, visual semantics, and flow are co-equal modalities of one diffusion transformer.
Following UWM~\cite{uwm} and UVA~\cite{uva}, we call the model \textbf{unified}, as one model that is trained jointly on the three modalities serves as a policy, as forward and inverse dynamics, and as a predictor of future world state.

Unlike the flow-based policies above, flow here is neither a separate module nor a tracked input, but a modality inside the model. From the same demonstration clips used for policy learning, we extract point maps and scene flow offline with a monocular 3D tracker~\cite{track4world} and encode them with a frozen 4D VAE~\cite{motioncrafter}. Each modality has its own expert, and the three exchange information only through shared joint attention. We add flow to a pretrained visual--action model during mid-training and post-training. The visual and action experts start from the pretrained weights, while the flow expert is trained from scratch. All three experts are then trained jointly, so the pretrained representations adapt to the new modality.

Each function of the model corresponds to one training objective. Ten objectives form a role grid, which assigns each modality the role of \emph{target}, \emph{condition}, or \emph{absent}. With one target per objective, the grid covers all four combinations of flow and future visual features being present or absent for action prediction, and retains three objectives each for flow and visual prediction. This deliberate asymmetry design prioritizes the conditioning choices for the control variable. The four objectives without flow recover the pretrained model's task set. Under the direct policy objective, the flow stream carries learned placeholders, so deployment requires no scene-flow extraction or flow encoding.

We evaluate UniFlowWAM in simulation on RoboCasa-GR1~\cite{robocasa}, LIBERO, and LIBERO-Plus, and on four real-world bimanual tasks. On RoboCasa-GR1, adding flow to the base model improves the average success rate by 4.3 percentage points. The gain is largest on the tasks where the base model is weakest (Sec.~\ref{sec:sim}). On LIBERO, UniFlowWAM preserves in-distribution success at 98.1\%, and it reaches 87.7\% zero-shot on LIBERO-Plus under out-of-distribution perturbations. On the real robot, UniFlowWAM achieves an average success rate of 87.0\%, compared with 68.5\% for $\pi_{0.5}$~\cite{pi05}, the strongest baseline (Sec.~\ref{sec:real}). Ablations show where the flow objectives matter. On the real robot, post-training with the full ten-objective grid outperforms policy-only post-training by 12 to 18 points on every task, and the gap is largest on object placements at the edge of the demonstrated region.

\noindent\textbf{Contributions.} (i)~We make scene-level 3D flow a co-equal modality of a unified visual--action model, added to a pretrained model through a dedicated expert and joint mid-training and post-training on the same demonstrations. (ii)~We organize actions, 3D flow, and future visual features through ten role-conditioned objectives, allowing flow to serve as a target or a condition while retaining a direct policy that needs no flow input. (iii)~We show manipulation gains on RoboCasa-GR1 and improved robustness on LIBERO-Plus while retaining in-distribution performance on LIBERO.

\section{Related Work}

\noindent\textbf{Unified models for manipulation.} Most robot foundation models scale behavior cloning over expert demonstrations~\cite{pi0,gr00t}. Some other models also learn to predict the future. VPP~\cite{vpp} conditions actions on the predicted future representations of a video diffusion model, and unified models such as UWM~\cite{uwm}, UVA~\cite{uva}, and Motus~\cite{motus} predict future observations jointly with actions. Recent work, including DINO-WM~\cite{dinowm}, FLARE~\cite{flare}, and LDA~\cite{lda}, predicts the features of a pretrained visual encoder instead of pixels. UniVLA~\cite{univla} learns latent actions in such a feature space, so that videos without action labels can be used to pretrain the policy. SpatialVLA~\cite{spatialvla}, MolmoAct~\cite{molmoact}, DreamVLA~\cite{dreamvla}, and X-WAM~\cite{xwam} further use depth to provide 3D information about the scene.

\noindent\textbf{From optical flow to scene-level 3D flow.} Flow was classically estimated in 2D, between adjacent frames~\cite{raft}. Such flow is limited to the pixel level, with no awareness of 3D geometry. Lifting flow to 3D requires per-pixel geometry, which feed-forward models such as VGGT~\cite{vggt} now recover from ordinary video. 3D point trackers such as SpatialTrackerV2~\cite{spatialtrackerv2} estimate world-space tracks jointly with depth and camera motion. Track4World~\cite{track4world} and TrackingWorld~\cite{trackingworld} extend such tracking to nearly all pixels, while MotionCrafter~\cite{motioncrafter} estimates dense geometry and 3D flow directly. Either route yields scene-level 3D flow from monocular video alone. These advances make 3D flow a practical training modality, rich in 3D information and cheap to obtain.

\noindent\textbf{How flow is used in policies.} In the most direct use, flow serves as an explicit interface between perception and control. NovaFlow~\cite{novaflow} extracts 3D object flow from generated videos without training a flow predictor, and plans actions from it. Im2Flow2Act~\cite{im2flow2act} and PointAction~\cite{pointaction} predict flow or point trajectories and convert them into actions with a separate module, so the predictor is trained without action labels and reused across embodiments. Flow can also be used inside a single policy. TraceVLA~\cite{tracevla}, MolmoAct~\cite{molmoact}, 3D Flow Diffusion Policy~\cite{fdp3d}, and LaMP~\cite{lamp} condition the actions on tracked or predicted flow at inference. Other methods use flow as a training signal. FlowVLA~\cite{flowvla} and Track4Action~\cite{track4action} use it as supervision and keep it out of policy inference. Concurrently, UMA~\cite{uma} and ChronoFlow-Policy~\cite{chronoflow} learn 3D keypoint trajectories jointly with actions.

\section{Method}
\begin{figure*}[t]
\centering
\includegraphics[width=0.95\textwidth]{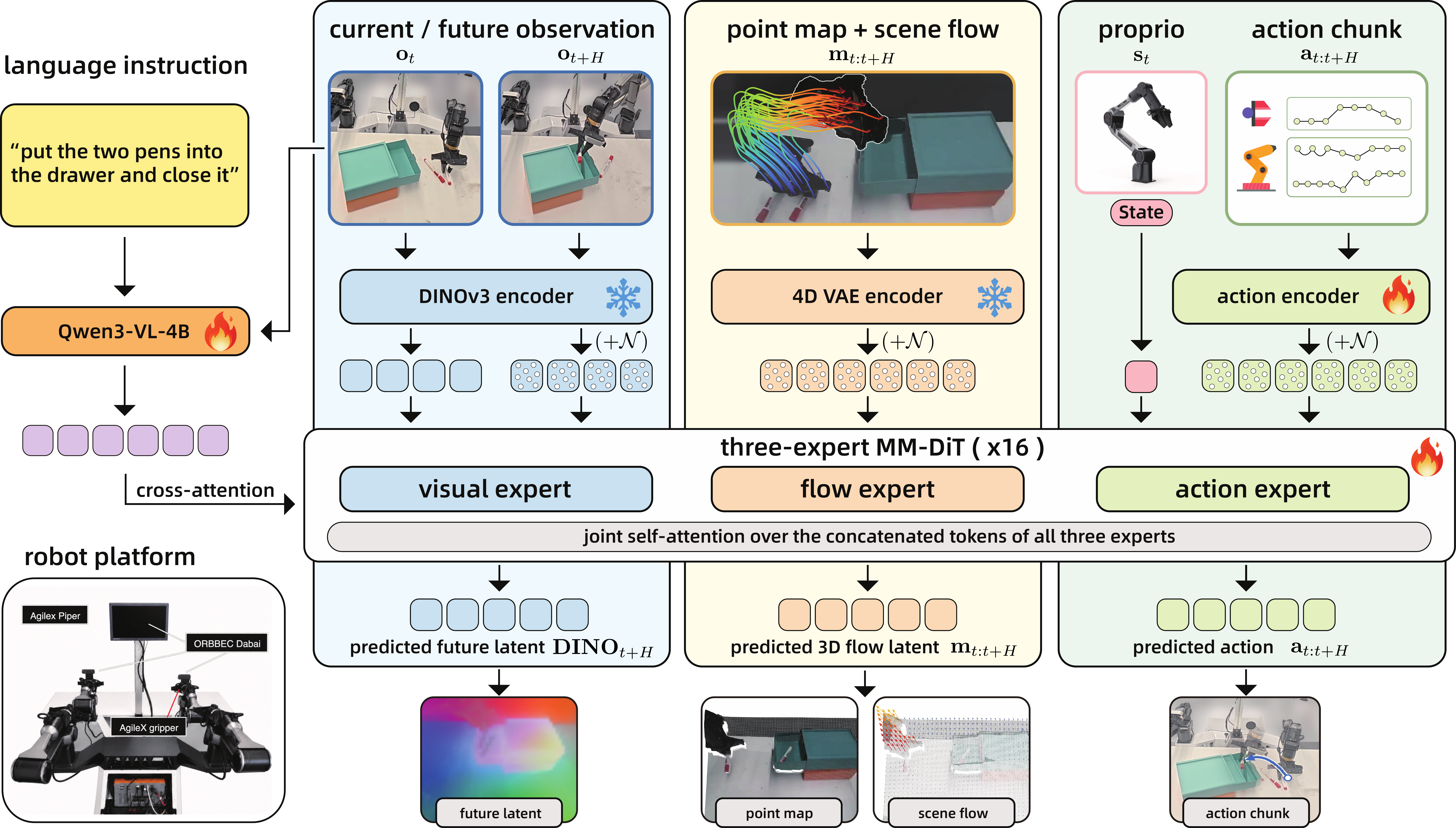}
\caption{\textbf{Network architecture.} Actions, scene-level 3D flow, and future visual features are three modalities of one diffusion transformer, each processed by its own expert; the experts exchange information only through joint self-attention. The objective decides which modalities are noised targets and which are clean conditions. Predicted flow latents are decoded by the 4D VAE into point maps and scene flow.}
{}
\label{fig:arch}
\end{figure*}

UniFlowWAM models robot actions, scene-level 3D flow, and future visual features within a single multimodal diffusion transformer (MMDiT)~\cite{mmdit} (Fig.~\ref{fig:arch}). Building on the visual--action backbone of LDA~\cite{lda}, we introduce a dedicated expert for the 3D flow representation and train the model with 10 role-conditioned objectives. Each objective specifies a prediction target and the modalities available as conditions.

\subsection{Problem Formulation}
\label{sec:prelim}

At time $t$, the model receives the context $c_t=(\mathbf{o}_t,\mathbf{s}_t,\ell)$: the visual observation, the proprioceptive state, and the language instruction. Three variables are defined over a horizon of $H$ frames. The action chunk $\mathbf{a}=\mathbf{a}_{t:t+H}$ is the command sequence, and the future visual feature $\mathbf{v}$ refers to the frozen DINOv3 feature~\cite{dinov3} of the frame $\mathbf{o}_{t+H}$ at the end of the horizon and serves as the visual-semantic part of the world representation. The 3D flow representation $\mathbf{m}$ summarizes how the scene moves over $[t,t{+}H]$ (Sec.~\ref{sec:flow}). Every objective in Sec.~\ref{sec:grid} is one conditional of the joint $p(\mathbf{a},\mathbf{m},\mathbf{v}\,|\,c_t)$.

\subsection{Scene-Level 3D Flow Representation}
\label{sec:flow}
The 3D flow is defined by the displacement of each point in the scene to the next frame and by the 3D position of that point, so that it also carries the structure of the scene. A pretrained dense 3D point tracker~\cite{track4world} produces this pair for the clip over $[t,t{+}H]$. We take $\mathbf{m}$ to be the latent of a pretrained 4D VAE~\cite{motioncrafter}, which jointly encodes both the point map of the scene and the 3D flow. 

For frame $i$ and pixel $u$, the tracker returns $P_i(u)$, the 3D point seen at that pixel, and $F_i(u)$, the displacement of that point to the next frame. Every frame is anchored on its own pixel grid, and no point is followed beyond the next frame. The 4D VAE uses two frozen encoders to separately encode the
3D position and displacement sequences, then concatenates their
outputs along the channel dimension,
$
\mathbf{m} = \big[
\mathcal{E}_{P}(P_{t:t+H});\,
\mathcal{E}_{F}(F_{t:t+H})
\big],
$
where the two components represent scene geometry and
frame-to-frame flow, respectively. Its decoder $\mathcal{D}$ recovers both $P$ and $F$ from $\mathbf{m}$, so a predicted $\mathbf{m}$ can be decoded and inspected. The pipeline is monocular. Flow is extracted from the same RGB videos used for training, so no depth sensor is needed for data collection or deployment.

\subsection{Three-Expert Setup}
\label{sec:arch}

\noindent\textbf{From variables to tokens.} Each of the three variables enters through its own mapping. 
Action tokens encode the action chunk and are prefixed by state tokens. 
A visual token carries two things at once: the DINOv3 features of the current frame, and the future-frame slot; a linear layer merges them, so the current observation is present under every objective. 
Flow arrives as the latent $\mathbf{m}$, already encoded offline by the 4D VAE. A \emph{flow patchifier} then cuts $\mathbf{m}$ into the tokens of the flow expert. The rest of the context enters as conditioning.
The instruction and the current frame are encoded by the Qwen3-VL-4B vision--language model (VLM), and each expert reads the VLM outputs through cross-attention.

\noindent\textbf{Within a modality, and across them.} Each expert keeps its own normalization, its own Q/K/V projections, its own feed-forward network and its own AdaLN modulation, so no parameter is shared between modalities. They meet in joint self-attention. Writing $\hat{x}^{k}$ for the modulated tokens of modality $k\in\{a,m,v\}$, each expert projects its own queries, keys and values, $(Q^{k},K^{k},V^{k})=W^{k}\hat{x}^{k}$, with private weights $W^{k}$. Its queries then attend over the keys and values of all three.

This is the one place where a modality sees the other two, and query--key normalization bounds its logits. Besides the learned absolute position embedding of the base model, added by sequence index, MM-RoPE~\cite{lumos} rotates the queries and keys of all three experts in one shared $(t,h,w)$ frame over the horizon and the visual patch grid. A visual token takes $t=H$, the frame it predicts, and its patch position. Action and state tokens take their step or frame index, with $(h,w)=(0,0)$. The flow patchifier merges two latent frames into each token, so the $t'$-th flow token takes $t=2t'$, and the center of its latent cell is rescaled onto the visual patch grid.

\noindent\textbf{From tokens back to variables.} Each expert ends in its own head, which predicts a flow-matching velocity~\cite{flowmatching} in that modality's own space, and therefore output DINO latent, Flow latent, or action latent.

\subsection{Role-Conditioned Multi-Objective Learning}
\label{sec:grid}
A visual--action model has four natural tasks. They come from choosing which of action and future visual features is the target and whether the other is given, namely policy $p(\mathbf{a}\,|\,c_t)$, forward dynamics $p(\mathbf{v}\,|\,\mathbf{a},c_t)$, inverse dynamics $p(\mathbf{a}\,|\,\mathbf{v},c_t)$, and visual prediction $p(\mathbf{v}\,|\,c_t)$~\cite{uwm}. When we add the third modality, it naturally extends the set in the same way. An objective assigns each of the three modalities one of three roles. \emph{Target} means the slot is noised and supervised. \emph{Condition} means it holds a clean input, with its flow-matching time fixed at the clean end. \emph{Absent} means the slot holds a learned placeholder
that remains visible to attention and indicates that the
modality is not provided for the current objective. Table~\ref{tab:grid} lists the ten objectives we use, in three kinds. Three predict one modality from the context alone (rows 1--3), namely policy, visual prediction, and flow prediction. Six predict one modality with one other given (rows 4--9). For each target, they cover the two ways of conditioning on a single other modality. These nine are symmetric across the three modalities. The tenth conditions the action on both flow and visual features (row 10). We add it for the action alone, as action is the highlighted modality for downstream use. 

The objective is a learned embedding, added to the timestep embedding before AdaLN. Each training sample draws one objective. A task-balanced sampler keeps every objective present in each batch, and the loader skips the modalities that objective does not need. The loss is the flow-matching velocity loss, on target slots only,
\begin{equation}
\mathcal{L}=\textstyle\sum_{k\in T(r)}\mathcal{L}^{(k)}_{\mathrm{FM}},\qquad k\in\{a,m,v\},
\label{eq:loss}
\end{equation}
where $r$ indexes the objective, $T(r)$ denotes its target set, and
$\mathcal{L}^{(k)}_{\mathrm{FM}}$ is the flow-matching loss for expert $k$. Condition slots receive clean inputs and no loss, and absent slots receive neither.

\begin{table}[t]
\centering
\caption{The ten training objectives (Sec.~\ref{sec:grid}).}
\label{tab:grid}
\footnotesize
\setlength{\tabcolsep}{3.2pt}
\begin{tabular}{llcccc}
\toprule
\# & Objective & Fixed &  Cond. & Abs. & Tar. \\
\midrule
1 & policy$^\dagger$     &      $c_t$       & --     & $\mathbf{m}, \mathbf{v}$ & $\mathbf{a}$ \\
2 & visual prediction$^\dagger$   &      $c_t$     & --     & $\mathbf{a}, \mathbf{m}$ & $\mathbf{v}$ \\
3 & flow prediction           &      $c_t$     & --     & $\mathbf{a}, \mathbf{v}$ & $\mathbf{m}$ \\
\midrule
4 & inverse dynamics$^\dagger$    &      $c_t$       & $\mathbf{v}$    & $\mathbf{m}$    & $\mathbf{a}$ \\
5 & flow-cond.\ policy            &      $c_t$       & $\mathbf{m}$    & $\mathbf{v}$    & $\mathbf{a}$ \\
6 & forward dynamics$^\dagger$    &      $c_t$       & $\mathbf{a}$    & $\mathbf{m}$    & $\mathbf{v}$ \\
7 & dynamics from flow            &      $c_t$       & $\mathbf{m}$    & $\mathbf{a}$    & $\mathbf{v}$ \\
8 & visual flow estimation         &      $c_t$      & $\mathbf{v}$    & $\mathbf{a}$    & $\mathbf{m}$ \\
9 & action-driven flow generation    &      $c_t$    & $\mathbf{a}$    & $\mathbf{v}$    & $\mathbf{m}$ \\
\midrule
10 & dual-cond.\ inverse dynamics     &      $c_t$    & $\mathbf{m}, \mathbf{v}$ & --     & $\mathbf{a}$ \\
\bottomrule
\end{tabular}
\tabnote{Fixed denotes the context $c_t$ shared by all objectives.
Cond. = Condition; Abs. = Absent; Tar. = Target.
$\mathbf{a}$, $\mathbf{m}$, and $\mathbf{v}$ denote action, 3D flow, and visual features,
respectively. A dash indicates no modality assigned to that role.
$\dagger$ marks the four tasks of the base model.}
{}
\end{table}

\subsection{Training Strategy}
\label{sec:training}
We initialize the visual--action backbone from the publicly released pretrained checkpoint of LDA~\cite{lda}. We then perform \emph{mid-training} to incorporate scene-level 3D flow into the model, jointly optimizing the ten objectives in Sec.~\ref{sec:grid} over actions, future visual features, and 3D flow. This stage extends the pretrained visual--action model to support 3D flow as both a prediction target and a conditioning modality. The resulting model is subsequently post-trained on demonstrations from each downstream domain.

\section{Experiments}
\label{sec:exp}

Our experiments assess manipulation performance, robustness to distribution shifts, and the contribution of scene-level 3D flow. We evaluate UniFlowWAM on RoboCasa-GR1, LIBERO, and LIBERO-Plus in simulation, and on an ALOHA-style robot for real-world bimanual manipulation.

\subsection{Experimental Setup}
\label{sec:exp_setup}

\noindent\textbf{Mid-training data.} We use a filtered mixture of three public robot-demonstration datasets; all are robot-camera recordings. RoboCOIN~\cite{robocoin} is a bimanual manipulation dataset collected on 15 robot platforms in residential, commercial, and industrial settings; we use 47{,}050 episodes from six of these platforms. RoboMIND~2.0~\cite{robomind2} is a dual-arm and mobile manipulation dataset collected on six robot embodiments; we use 140{,}231 episodes from three of them, including Franka and UR arms. MolmoAct2-BimanualYAM~\cite{molmoact2} is AllenAI's teleoperated bimanual dataset collected on a single low-cost platform, the YAM arms; we use 27{,}162 episodes. The combined corpus covers ten embodiments, with 214{,}443 episodes and 133.0 million frames. We sample 16-step chunks at three-frame intervals, yielding around 42 million training windows. We extract scene-level 3D flow offline with frozen Track4World~\cite{track4world} and encode its point maps and flow into latents using a frozen MotionCrafter 4D VAE~\cite{motioncrafter}.

\noindent\textbf{Model and training.} Starting from LDA's released pretrained checkpoint~\cite{lda}, our model UniFlowWAM adds a dedicated 3D flow expert to the MMDiT backbone to incorporate flow as a third modality. Qwen3-VL-4B supplies visual--language conditioning through cross-attention, while DINOv3 provides visual prediction targets. Both encoders remain frozen during mid-training. We train on 48 NVIDIA H200 GPUs with a global batch size of 1{,}536.

\subsection{Simulation Experiments}
\label{sec:sim}

\subsubsection{Benchmarks}
\label{sec:benchmarks}
RoboCasa~\cite{robocasa} targets everyday household tasks for generalist robots. We use its GR-1 tabletop tasks: 24 tasks for a GR-1 humanoid with dexterous hands, observed by a single egocentric camera and controlled through 29 joint dimensions across the arms, hands, and waist. Each task comes with 1{,}000 demonstrations, 24{,}000 in total, or 6.0 million frames and 83.6 hours at 20 fps.

LIBERO~\cite{libero} targets lifelong robot learning with a single Franka Panda arm. It contains 4 suites of 10 tasks each: Spatial varies the layout of the same objects, Object varies the objects to manipulate, Goal varies the goal within one scene, and Long collects long-horizon tasks. We use the no-noops release introduced by OpenVLA~\cite{openvla}. We evaluate in two settings. LIBERO itself is the in-distribution test. LIBERO-Plus~\cite{liberoplus} perturbs the same tasks along seven axes, camera viewpoint, robot initial state, language, lighting, background, sensor noise, and object layout, for 10{,}030 episodes in total. Following its official zero-shot protocol, we train only on LIBERO and evaluate on LIBERO-Plus.

\subsubsection{Post-training and Results}
\label{sec:posttraining}
On RoboCasa-GR1, we train one model on all 24 tasks jointly. UniFlowWAM without mid-training starts from the released LDA pretrained checkpoint, and UniFlowWAM with mid-training starts from our mid-training checkpoint. Both follow a 500k-step cosine schedule with 5k warmup steps. The learning rate is $10^{-4}$ for the action model and $10^{-5}$ for the VLM, while DINOv3 remains frozen.

\begin{table}[t]
\centering
\caption{Success rates (\%) on RoboCasa-GR1 tabletop.}
\label{tab:robocasa}
\footnotesize
\setlength{\tabcolsep}{6pt}
\begin{tabular}{lc}
\toprule
\multicolumn{2}{l}{\emph{Comparison with published results}} \\
Model & SR (\%) $\uparrow$ \\
\midrule
UWM~\cite{uwm} & 14.2 \\
GR00T-N1.6~\cite{gr00t} & 47.6 \\
StarVLA~\cite{starvla} & 47.8 \\
Being-H0.7~\cite{beingh07} & 49.2 \\
FLARE~\cite{flare} & 55.0 \\
LDA-1B~\cite{lda} & 55.4 \\
UniFlowWAM & \underline{58.9}$^{\dagger}$ \\
OpenWAM-$\alpha$~\cite{openwam} & \textbf{60.5} \\
\midrule
\multicolumn{2}{l}{\emph{Repeated evaluation (seed 0--seed 10)}} \\
Model & Average $\pm$ SD (\%) $\uparrow$ \\
\midrule
LDA (released) & 52.11 $\pm$ 2.24 \\
LDA (reproduced) & 52.76 $\pm$ 1.16 \\
UniFlowWAM w/o mid-training & \underline{54.84 $\pm$ 1.00} \\
UniFlowWAM w/ mid-training & \textbf{56.38 $\pm$ 1.47} \\
\bottomrule
\end{tabular}
\tabnote{$\dagger$: UniFlowWAM's best result over evaluation seeds 0--10. Bold marks the best score in each block, underline the second best.}
{}
\end{table}

We evaluate released and reproduced LDA checkpoints and both UniFlowWAM variants over 11 shared evaluation seeds (0--10), with 50 episodes per task and seed. UniFlowWAM achieves mean SRs of 54.84\% without mid-training and 56.38\% with mid-training (Table~\ref{tab:robocasa}, bottom). Both exceed released LDA (52.11\%) and reproduced LDA (52.76\%). UniFlowWAM with mid-training improves over released LDA by 4.27 percentage points. The gain concentrates on the tasks where released LDA is weaker. On the 10 tasks where its mean SR is below 50\%, UniFlowWAM with mid-training improves on all 10, by 6.47 percentage points on average; on the remaining 14 tasks the average gain is 2.71 points.

\begin{table}[t]
\centering
\caption{Success rates (\%) on LIBERO and LIBERO-Plus.}
\label{tab:libero}
\label{tab:liberoplus}
\scriptsize
\setlength{\tabcolsep}{0.8pt}
\begin{tabular*}{\columnwidth}{@{\extracolsep{\fill}}lc|cccccccc@{}}
\toprule
Model & LIBERO & \multicolumn{8}{c}{LIBERO-Plus} \\
\cline{3-10}
 & Avg & Cam. & Robot & Lang. & Light & Bg. & Noise & Layout & Total \\
\midrule
OpenVLA-OFT~\cite{openvlaoft} & 97.1 & 56.4 & 31.9 & 79.5 & 88.7 & 93.3 & 75.8 & 74.2 & 69.6 \\
$\pi_{0.5}$~\cite{pi05} & 96.9 & 72.2 & \textbf{76.8} & 85.1 & 96.2 & 95.5 & 87.0 & \textbf{86.9} & \underline{84.7} \\
StarVLA-$\alpha$~\cite{starvlaalpha} & -- & 48.7 & 63.4 & 86.8 & 95.8 & 94.6 & 75.0 & 80.2 & 77.8 \\
Fast-WAM~\cite{fastwam} & 97.6 & 40.0 & 65.3 & 68.8 & 91.4 & 58.2 & 63.8 & 75.5 & 65.3 \\
4D-WAM~\cite{fourdwam} & \underline{98.6} & 45.2 & 64.3 & \textbf{90.6} & 94.3 & 57.7 & 69.1 & 79.2 & 71.0 \\
FLEX-$\pi$~\cite{flexpi} & 98.5 & 73.7 & 53.1 & 84.3 & 91.9 & 93.8 & \underline{93.4} & \underline{82.9} & 80.9 \\
OpenWAM-$\alpha$~\cite{openwam} & \textbf{99.3} & 33.8 & \underline{76.1} & 88.0 & \underline{97.0} & 87.1 & 39.8 & 77.5 & 69.2 \\
LaMP~\cite{lamp} & 98.3 & 64.5 & 69.6 & \underline{88.2} & 95.3 & \textbf{97.4} & 76.9 & 73.8 & 79.3 \\
Track4Action~\cite{track4action} & 97.0 & \underline{76.6} & 60.7 & 87.1 & 96.3 & \underline{96.1} & 88.4 & 78.7 & 82.3 \\
LDA~\cite{lda} & 92.8 & 33.1 & 71.2 & 76.0 & 89.9 & 91.1 & 72.0 & 66.2 & 69.5 \\
\midrule
UniFlowWAM (ours) & 98.1 & \textbf{96.9} & 66.0 & 87.4 & \textbf{98.0} & 87.3 & \textbf{98.1} & 81.7 & \textbf{87.7} \\
\bottomrule
\end{tabular*}
\tabnote{Bold marks the best score in each column, underline the second best. --: not reported.}
{}
\end{table}

On LIBERO, we post-train one model on the four suites jointly, for 120k steps by default, and evaluate it on LIBERO and, zero-shot without adaptation, on LIBERO-Plus. On LIBERO, UniFlowWAM reaches 98.05\% average SR (Table~\ref{tab:libero}). Leading methods already score 97--99\% on this benchmark; at this level of success rate observed, a difference of one percentage point is not informative, and we report LIBERO only to verify that in-distribution performance is preserved. LIBERO-Plus separates the methods more clearly. UniFlowWAM reaches 87.65\% with its best evaluated checkpoint, and Table~\ref{tab:liberoplus} reports its success rates across all seven perturbation categories. The margin is widest under camera perturbations, where UniFlowWAM reaches 96.9\%, against 76.6\% for the strongest baseline in that category.

\subsection{Real-World Experiments on ALOHA}
\label{sec:real}
\noindent\textbf{Hardware.} We use AgileX Cobot Magic, an ALOHA-style platform with two 6-DoF PiPER arms and parallel-jaw grippers. A top-mounted camera provides the main view, and two wrist cameras provide the wrist views. Both the proprioceptive state and the action command are 14-dimensional, with six joint positions and one gripper position per arm.

\begin{figure}[t]
\centering
\includegraphics[width=\figw]{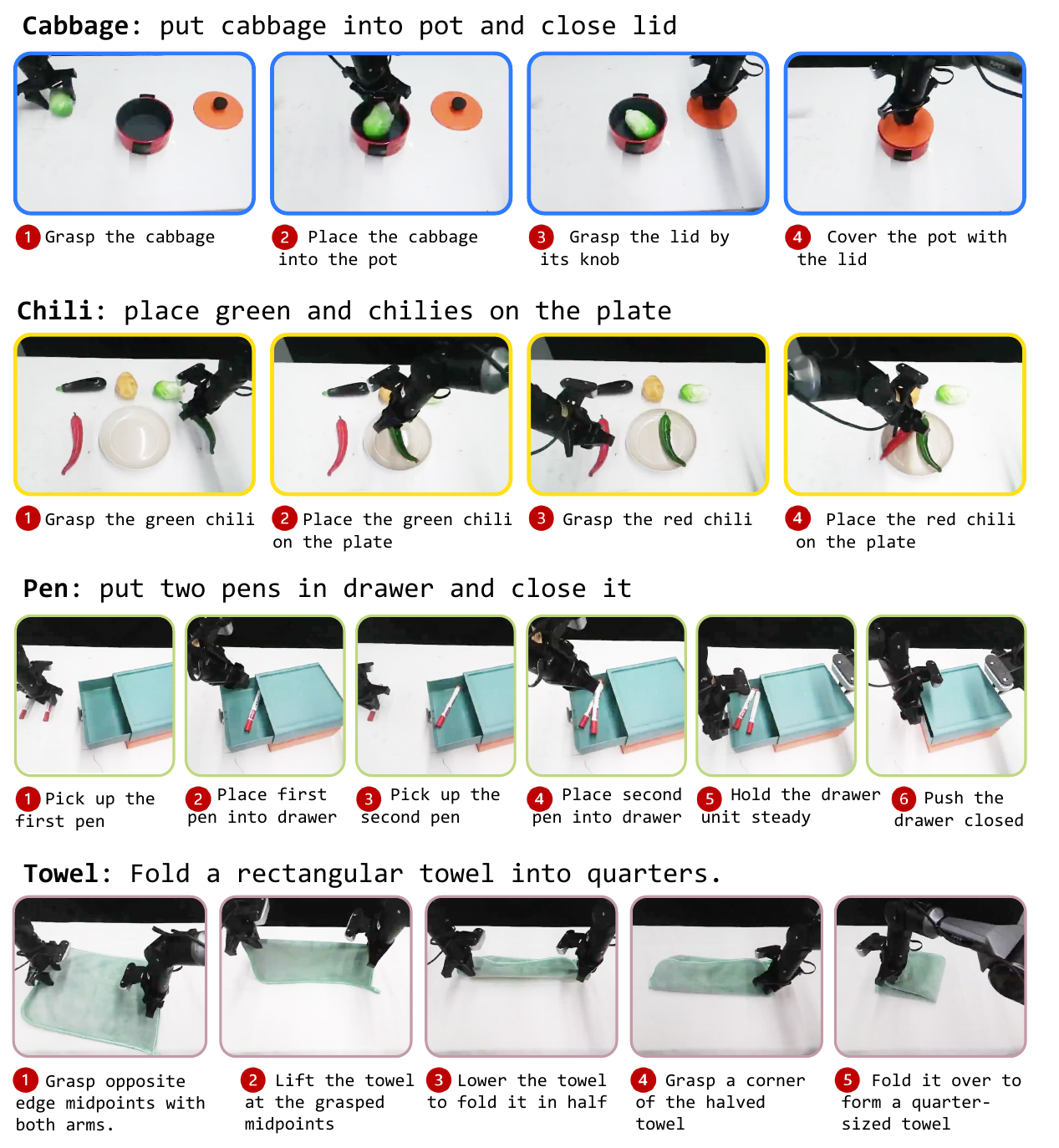}
{}
\caption{\textbf{Real-world manipulation tasks.}
Instructions and scored milestones for four tasks:
placing a cabbage in a pot and closing the lid,
transferring two chilies onto a plate,
placing two pens in a drawer and closing it,
and folding a towel.}
{}
\label{fig:real}
\end{figure}

\noindent\textbf{Demonstrations.} We consider four tasks that differ in what they demand of the two arms (Fig.~\ref{fig:real}). \emph{Chili transfer} repeats a single pick-and-place primitive, moving two chilies onto a plate. \emph{Cabbage transfer} chains distinct primitives instead of repeating one, placing a cabbage into a pot and then covering the pot with a lid. \emph{Pen manipulation} adds an articulated object and asymmetric arm roles, as one arm holds the drawer steady while the other places two pens into it in succession, after which the drawer is pushed closed. \emph{Towel folding} replaces the rigid objects with a deformable one that has no pose to track, requiring two successive folds. After preprocessing, we retain 50 demonstrations per task and post-train a separate model for each task. At inference, each model uses the policy mode to predict 16-step action chunks.

\noindent\textbf{Evaluation.} For model with full objective setup as our default model, each policy is evaluated for 50 trials per task. We fix the starting conditions in advance instead of drawing them at random. Within a task we change one thing at a time, in equal steps. For chili transfer we keep the chili at the same spot on the table and turn it about the vertical axis to five fixed orientations. Each orientation is run the same number of times, and the plate position varies slightly across trials. The other three tasks follow the same idea along the factor that matters for them. All conditions stay inside the area the demonstrations cover. Every policy sees the same starting conditions, so the differences we report come from the policies and not from where the objects happened to start. We report full-task success rate and a process score, the fraction of milestones completed, scaled to 100. The milestones are those in Fig.~\ref{fig:real}, four for cabbage (grasp, place in pot, grasp lid, close) and chili (grasp and place each chili), six for pen (pick and place each pen, support the drawer with the right gripper, push it closed), and five for towel (grasp, lift, fold in half, fold again, press flat). Milestones are scored independently, so a trial that fails one but recovers on a later one still earns credit for the later one. Both metrics are averaged over trials, then over tasks.

\begin{figure}[t]
\centering
\includegraphics[width=\figw]{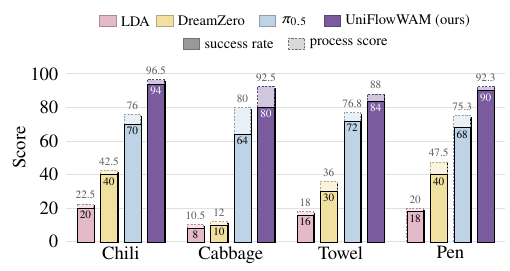}
{}
\caption{\textbf{Results of real-world manipulation on ALOHA.} Solid bars give the full-task success rate; the
lighter dashed segment extends to the process score, the fraction of ordered milestones
completed (Fig.~\ref{fig:real}), scaled to 100. 50 trials per task and policy.}
\label{fig:aloha}
\end{figure}

\noindent\textbf{Task performance.} UniFlowWAM reaches 87.0\% average success with a process score of 92.3, against 68.5\% and 77.0 for $\pi_{0.5}$, 30.0\% and 34.5 for DreamZero, and 15.5\% and 17.8 for LDA (Fig.~\ref{fig:aloha}). It is the strongest policy on every task. The two metrics also separate the failure modes. Averaged over the trials it does not complete, UniFlowWAM still earns 41 of 100 milestone points and $\pi_{0.5}$ earns 27, whereas LDA earns 3, so LDA tends to fail at the first milestone rather than partway through. The failure for LDA comes from severe arm tremor. Our reproduced LDA shakes strongly during execution, so the arm often knocks the object away before the gripper closes, and the trial rarely recovers. The two metrics diverge most on cabbage transfer. Its first two milestones handle the cabbage and the last two handle the lid. The round cabbage can roll out of the gripper while the lid is still grasped and closed, so UniFlowWAM succeeds on 80\% of trials yet completes 92.5\% of the milestones.

\subsection{Ablation Studies}
\label{sec:abl}
\begin{figure}[t]
\centering
\newcommand{\coveragepanelheight}{90bp}
\input{sections/figs/chili_orientation_plot}\figpanelsep
\includegraphics[height=\coveragepanelheight]{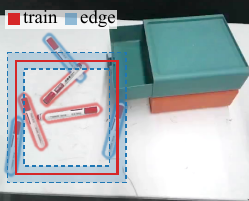}
{}
\caption{\textbf{Demonstration coverage on two real-robot tasks.}
\textbf{Left:} chili transfer, with coverage measured over the chili's rotation
angle: 0$^\circ$ is lengthwise, $\pm$90$^\circ$ sideways. \textbf{Right:} pen
manipulation, with coverage measured over the pen's placement in the top camera.
The red box spans the initial pen positions of all 49 demonstrations; the blue
band is its boundary, where demonstrations are naturally sparse.}
\label{fig:coverage}
\end{figure}

\noindent\textbf{Post-training objectives.} We first ask what the objective grid contributes on its own, varying the objectives used in LIBERO post-training while holding the mid-trained initialization and the demonstration data fixed. The four variants form a nested ladder. The full grid uses all ten objectives, with 3D flow as both a target and a condition. The flow-as-condition variant drops the three objectives that predict flow and keeps the other seven. The no-flow variant also drops flow conditioning and keeps the four visual--action objectives of the base model. Policy-only training keeps the single action-prediction objective. Every variant acts through the direct policy objective and takes no flow input at test time, so the differences come from training rather than from the inference path.

All four objective sets are first post-trained on the same 120k-step schedule. The reduced sets level off within this budget, whereas the full ten-objective grid is still improving at 120k steps. We therefore extend the full grid alone to a 200k-step schedule, and at 170k steps it reaches 87.0\% on LIBERO-Plus. More objectives thus call for a longer schedule rather than lowering the attainable success rate. Because the simulated benchmark separates the objective sets mainly by training speed, we compare them on the real robot below, where at equal training steps the full grid clearly outperforms policy-only training.

\begin{table}[t]
\centering
\caption{Objective ablations on ALOHA (success rate, \%).}
\label{tab:aloha_objectives}
\footnotesize
\setlength{\tabcolsep}{3pt}
\begin{tabular}{@{}p{0.52\columnwidth}rrr@{}}
\toprule
Training objectives & Chili & Pen & Towel \\
\midrule
obj-1: policy only & 76 & 76 & 72 \\
obj-4: without flow objectives & 92 & 80 & 76 \\
obj-7: flow as condition only & 92 & 88 & 76 \\
\midrule
obj-10: full UniFlowWAM & \textbf{94} & \textbf{90} & \textbf{84} \\
\bottomrule
\end{tabular}
\tabnote{The three ablation rows are measured over 25 trials per task. The obj-10 row below the rule is the main experiment, measured over 50 trials, and is the run reported in Fig.~\ref{fig:aloha}. The obj-4 variant retains the four visual--action objectives and omits all 3D flow objectives. The obj-7 variant adds three objectives that use 3D flow as a condition but never as a prediction target. These variants change the post-training objectives while retaining the model architecture, and all of them deploy without flow input.}
\end{table}

\noindent\textbf{Objectives on the real robot.} On the real robot, the four variants differ noticeably in success rate (Table~\ref{tab:aloha_objectives}). Policy-only post-training reaches 76\% on chili transfer, 76\% on pen manipulation, and 72\% on towel folding, while the full ten-objective grid reaches 94\%, 90\%, and 84\%, a gain of 12 to 18 points on every task. The no-flow variant, which keeps the four visual--action objectives, closes most of this gap on chili transfer (92\%), but the gap largely remains on pen manipulation (80\%), where the two variants that use 3D flow reach 88\% and 90\%. On all three tasks, success rate improves overall as objectives are added. Relative to the no-flow variant, the 3D flow objectives add 10 points on pen manipulation and 8 on towel folding, the two more demanding tasks, but only 2 on chili transfer, the simplest one.

\noindent\textbf{Where the failures fall.} The losses concentrate in one place. Chili transfer splits its trials evenly into five orientation groups, and 4 of the 6 failures of policy-only post-training fall in the sideways $90^\circ$ group alone. In that group, policy-only post-training succeeds in only 1 of its 5 trials, a success rate of 20\%, against 76\% over the whole task. The full grid succeeds in 80\% of its trials in the same group. Although the group contains few trials, a gap of 20\% against 80\% is large enough that we read it as a real difference rather than as sampling noise.

Why that group is hard is visible in the training data. Fig.~\ref{fig:coverage} shows that the demonstrations are far from uniform. Most chili grasps cluster near the lengthwise placement, so the sideways orientation is the condition the model has seen least, and it is where policy-only post-training fails most. In the pen demonstrations, most pens start in the middle of a compact region, and few start near its edge. When the pens start near this edge, policy-only post-training succeeds less often.

The two variants differ in one respect. Policy-only post-training never sees 3D flow, neither as a conditioning input nor as a prediction target, whereas the full grid uses it as both. Initialization, data, schedule, and the deployment path are identical, and neither takes flow as input at test time, so the difference at the edge of the demonstrated region is attributable to the flow objectives themselves.

\noindent\textbf{Position encoding.} Removing MM-RoPE lowers LIBERO-Plus at every evaluated checkpoint, by 1.1--2.8 points and by 2.5 at the best checkpoint, with the loss concentrated under robot-initial-state perturbations. Removing the absolute embedding costs 1.5 points at the best checkpoint and up to 4.4 at matched steps, with the gap narrowing as training proceeds. We keep both. Observation and flow tokens lie on the image grid, while action tokens have no image location. We tested two placements for the action tokens, the grid origin and the grid center. The origin is better by less than one point in the best LIBERO success rate. We therefore adopt the grid origin, while considering the difference between the two placements to be small.

\subsection{Discarded Designs}
\label{sec:discarded}
While developing joint training with 3D flow, we tried several designs that did not work well. The two below were not run as controlled ablations, but each bears on a natural design choice.

\noindent\textbf{Encoding flow with an image VAE.} Im2Flow2Act~\cite{im2flow2act} wrote 2D point tracks into image channels and encoded them with a Stable Diffusion autoencoder. Our first version extended this idea to 3D. It mapped the displacement $(\Delta x, \Delta y, \Delta z)$ of each tracked point, relative to its position in the first frame of the clip, directly to the R, G, and B channels and encoded the result with the frozen SDXL image VAE. The VAE reconstructs these images faithfully, and the flow denoising loss decreases normally. The generated flow, however, is nearly the same for every input, so the model has collapsed to an average flow over the dataset. Revising the training data and the patch embedding reduced the collapse but never produced usable flow. We later replaced the image VAE with the 4D VAE of MotionCrafter~\cite{motioncrafter}, which is trained on geometry and flow.

\noindent\textbf{Gating the flow modality.} To preserve what the LDA backbone learned in pretraining, we first admitted flow gradually through a per-layer gate on the attention logits from the existing tokens to the flow tokens. The gate starts closed and opens on a fixed fast or slow schedule. On RoboCasa-GR1, neither schedule helps. The faster one trails the ungated model by 3.3 points in mean success rate over 21 seeds. All later experiments, including every UniFlowWAM result in this paper, therefore use no gate.

\section{Limitations}
\noindent\textbf{Flow tracking noise and representation shift.} We use Track4World to generate offline 3D flow pseudo-labels from demonstration videos. These labels can contain temporal inconsistencies, such as an isolated reversal of flow direction within a continuous trajectory. In our pipeline, we use SAM-based segmentation to separate the manipulation foreground from the background and retain only foreground 3D flow for latent encoding. The pretrained 4D VAE~\cite{motioncrafter} remains effective at compressing these foreground-only inputs.

\noindent\textbf{Limited flow pretraining.} Our 3D flow mid-training uses approximately 1{,}200 hours of demonstrations, substantially less than the EI-30k corpus used for LDA pretraining~\cite{lda}. The added flow expert has not been pretrained alongside the visual--action backbone on that scale. 
 As tracking models continue to improve, flow extraction is expected to become faster and more accurate, making flow easier to scale for robot training in the same way as video. We leave larger-scale joint pretraining to future work.

\bibliography{references}

\end{document}